\documentclass[letterpaper]{article} % DO NOT CHANGE THIS
\usepackage{aaai2026}  % DO NOT CHANGE THIS
\usepackage{times}  % DO NOT CHANGE THIS
\usepackage{helvet}  % DO NOT CHANGE THIS
\usepackage{courier}  % DO NOT CHANGE THIS
\usepackage[hyphens]{url}  % DO NOT CHANGE THIS
\usepackage{graphicx} % DO NOT CHANGE THIS
\usepackage{natbib}  % DO NOT CHANGE THIS AND DO NOT ADD ANY OPTIONS TO IT
\usepackage{caption} % DO NOT CHANGE THIS AND DO NOT ADD ANY OPTIONS TO IT
\usepackage{algorithm}
\usepackage{algorithmic}
\usepackage{amsmath}
\usepackage{newfloat}
\usepackage{listings}
\usepackage{multirow}
\usepackage{booktabs}  % 用于三线表
\usepackage{subfigure}
\usepackage{pifont}  % 放在导言区

\title{BAPFL: Exploring Backdoor Attacks Against \\Prototype-based Federated Learning}
\author{
    Honghong Zeng\textsuperscript{\rm 1},
    Jiong Lou\textsuperscript{\rm 1,2},
    Zhe Wang\textsuperscript{\rm 1},
    Hefeng Zhou\textsuperscript{\rm 1},
    Chentao Wu\textsuperscript{\rm 1,2},
    Wei Zhao\textsuperscript{\rm 3},
    Jie Li\textsuperscript{\rm 1,2}\thanks{Jie Li is the corresponding author.}
}
\affiliations{
    \textsuperscript{\rm 1}School of Computer Science, Shanghai Jiao Tong University\\
    \textsuperscript{\rm 2}Yancheng Blockchain Research Institute, Shanghai Jiao Tong University (Wuxi) Blockchain Advanced Research Center, Shanghai Key Laboratory of Trusted Data Circulation and Governance, and Web3\\
    \textsuperscript{\rm 3}Shenzhen University of Advanced Technology, Shenzhen Institute of Advanced Technology, Chinese Academy of Science\\
    \{zhhhhh5, lj1994, zhe.wang, zadeji1\}@sjtu.edu.cn, wuct@cs.sjtu.edu.cn, weizhao86@outlook.com, lijiecs@sjtu.edu.cn
}

\usepackage{bibentry}
\begin{document}

\maketitle

\begin{abstract}

Prototype-based federated learning (PFL) has emerged as a promising paradigm to address data heterogeneity problems in federated learning, as it leverages mean feature vectors as prototypes to enhance model generalization. However, its robustness against backdoor attacks remains largely unexplored. In this paper, we identify that PFL is inherently resistant to existing backdoor attacks due to its unique prototype learning mechanism and local data heterogeneity. To further explore the security of PFL, we propose BAPFL, the first backdoor attack method specifically designed for PFL frameworks. BAPFL integrates a prototype poisoning strategy with a trigger optimization mechanism. The prototype poisoning strategy manipulates the trajectories of global prototypes to mislead the prototype training of benign clients, pushing their local prototypes of clean samples away from the prototypes of trigger-embedded samples. Meanwhile, the trigger optimization mechanism learns a unique and stealthy trigger for each potential target label, and guides the prototypes of trigger-embedded samples to align closely with the global prototype of the target label. Experimental results across multiple datasets and PFL variants demonstrate that BAPFL achieves a 35\%-75\% improvement in attack success rate compared to traditional backdoor attacks, while preserving main task accuracy. These results highlight the effectiveness, stealthiness, and adaptability of BAPFL in PFL.
\end{abstract}



\section{Introduction}
%data heterogeneous of FL
Federated learning (FL) is a distributed machine learning paradigm that enables multiple clients to collaboratively train a global model without sharing their private data, thus preserving data privacy. Due to this advantage, FL has been widely applied in various real-world scenarios, such as personalized recommendation \cite{fl-recommendation}, autonomous driving \cite{fl-driving}, and smart healthcare \cite{fl-medical}. In such practical applications, however, clients usually gather data from diverse sources, resulting in significant data heterogeneity. This data heterogeneity makes it challenging for a unified global model to achieve high performance on all clients. To address this challenge, many studies have focused on heterogeneous FL \cite{PFedBA,Cluster1,tan2021fedproto}. Among these approaches, prototype-based federated learning (PFL) \cite{tan2021fedproto} has shown great promise due to its ability to learn high-quality personalized models for clients with minimal communication overhead. %This makes it a potentially effective solution for heterogeneous FL systems.

%For instance, service providers of smart wearable medical devices can utilize FL to collaboratively optimize health monitoring models across different users. 

%The users’ physical conditions, data acquisition frequency, and device usage habits are different, which will lead to significant discrepancies in local data distributions across clients. It is a challenge for a single unified global model to achieve consistently high performance on all clients, and the personalization demands of different users are often unmet. 

%In such practical applications, however, data heterogeneity is a pervasive issue. . Due to the differences in users’ physical conditions, data acquisition frequencies, and device usage habits, the local data distributions across clients are significantly various. 
 
%Federated learning (FL) is a distributed machine learning framework that allows multiple clients to collaboratively train a global model by sharing local models instead of their private data to protect privacy. However, the trained global model of FL usually does not perform well on all clients when the data of clients is heterogeneous. For example, the service provider of smart wearable medical devices can use FL to optimize clients' health monitoring model. However, due to individual differences in physical conditions, living habits, and device usage frequency, the local data distribution of different clients may be significantly different. This heterogeneity making a unified global model difficult to ensure consistently high accuracy across all clients. 

\begin{figure}[t]
	\centering
	\includegraphics[width=\linewidth]{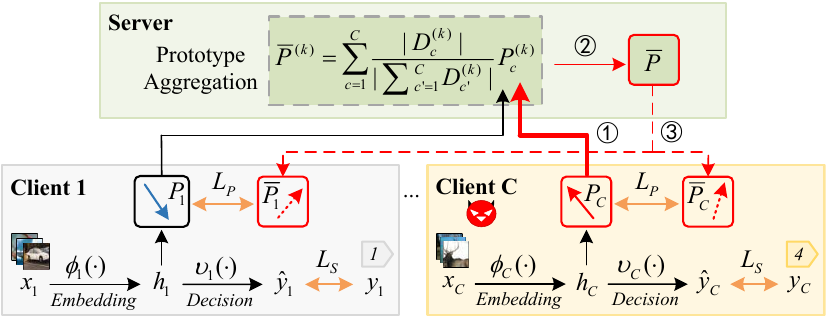}
	\caption{PFL system: one server and $C$ clients. The attacker manipulates client $C$ to upload poisoned prototypes $P_C$ (\ding{172}). $P_C$ deviates from the benign prototypes like $P_1$, thus it poisons the global prototypes $\Bar{P}$ (\ding{173}) and misleads the local training of benign models (\ding{174}).}\label{fig:pfl}
\end{figure}

%introduce PFL  
Unlike vanilla FL methods that aggregate full model parameters across clients, PFL~\cite{tan2021fedproto} exchanges \textit{class prototypes}, i.e., the average feature vectors of samples within the same class, to train models for clients. Typically, each client periodically updates its local prototypes and model by minimizing classification loss and aligning local prototypes with the global prototypes. The server then averages these local prototypes by class to form new global prototypes. Compared to FL, PFL significantly reduces communication overhead and improves model generalization under heterogeneous data \cite{tan2022federated, TAN2025FedPD}. With ongoing innovations in prototype representation~\cite{tan2022federated, proto-rep2,proto-rep3}, optimization objective~\cite{FedPLVM}, and robust aggregation~\cite{TAN2025FedPD,proto-agg}, PFL is expected to play a key role in real-world heterogeneous FL systems.

Despite its potential, the security of PFL remains underexplored. This research gap creates a critical blind spot: as illustrated in Figure \ref{fig:pfl}, attackers can exploit the prototype-sharing mechanism of PFL by manipulating some clients to upload poisoned prototypes (step \ding{172}). These prototypes deviate from benign prototypes uploaded by benign clients and cause the aggregated prototypes to drift from the correct direction (step \ding{173}), thereby misleading the local training of multiple benign clients (step \ding{174}). Affected by such attacks, the PFL system may cause serious consequences in security-critical applications such as medical diagnosis and financial decision-making.

% Therefore, it is imperative to study the security of PFL framework to uncover and mitigate its hidden vulnerabilities.

% Therefore, it is imperative to study backdoor attacks tailored to the PFL framework to uncover and mitigate its hidden vulnerabilities.
 %on federated learning
Among various federated attack strategies, backdoor attacks pose a particularly insidious and dangerous threat~\cite{SADBA2025}. These attacks inject poisoned samples with specific triggers into the training data to manipulate the model’s predictions. In this paper, we investigate the susceptibility of PFL frameworks to backdoor attacks. We first explore whether the PFL approach is still vulnerable to existing backdoor attacks. We observe that PFL exhibits strong robustness against existing backdoor attacks (see Section “Modeling and Analysis” for details). We attribute this robustness to two key factors: 
1) The limited influence of poisoned prototypes. Even if the global prototypes are contaminated by poisoned prototypes, they only affect the embedding layer of benign models. While the unaffected decision layer of the benign model obstructs the attack effectiveness. 2) Data heterogeneity of clients. Some clients may lack the training samples of the target label. Thus, their decision layer does not learn parameters for the target label. This inherently breaks the \textit{trigger-target label} mapping and significantly reduces the attack success rate (ASR). 

\begin{figure}[t]
	\centering
	\includegraphics[width=0.9\linewidth]{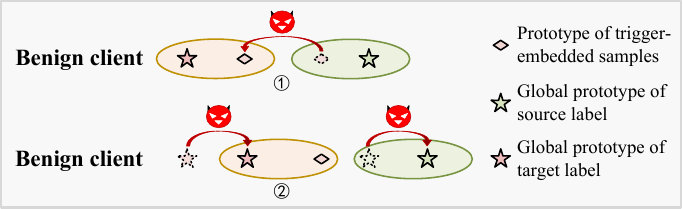}
	\caption{\ding{172}  Traditional backdoor attack strategy versus \ding{173} our backdoor attack strategy.}\label{fig:attack-idea}
\end{figure}

These factors motivate us to rethink backdoor attack strategies for PFL. As illustrated in Figure~\ref{fig:attack-idea}, while traditional backdoor attack can directly manipulate the trigger-embedded samples' classification in traditional FL by sharing full model parameters, we must strategically manipulate the global prototype to mislead the trigger-embedded samples' classification in an indirect manner. According to this analysis, we propose BAPFL, a novel backdoor attack method designed for PFL. BAPFL effectively attacks PFL systems from the perspective of dual-direction prototype optimization. Specifically, BAPFL comprises two components: 1) A prototype poisoning strategy (PPS) that leverages poisoned prototypes to manipulate the global prototype away from the prototypes of trigger-embedded samples (termed trigger prototypes), thereby guiding benign prototypes away from these trigger prototypes. 2) A trigger optimization mechanism (TOM) that ensures the attack’s effectiveness across heterogeneous clients. It learns stealthy triggers for dynamic target labels, and optimizes the trigger prototypes to closely align with the global prototype of the target label. These two modules jointly enhance the effectiveness of BAPFL, achieving high ASR and main task accuracy (ACC) across diverse PFL frameworks. Our main contributions are summarized as follows. 
%BAPFL systematically examines the feasibility of backdoor attacks in PFL from the perspective of dual-direction prototype optimization. 

\begin{itemize}
    \item This study delves into the security domain of PFL, and reveals that PFL exhibits strong resistance to conventional backdoor attacks. We identify two key factors behind this resistance: the limited influence of poisoned prototypes and data heterogeneity of clients. 
    
    \item We propose a novel backdoor attack method for PFL, called BAPFL, which combines PPS and TOM. PPS pushes benign prototypes away from trigger prototypes by manipulating global prototype aggregation, while TOM pulls trigger prototypes closer to the global prototypes of target labels by learning diverse stealthy triggers. This dual-direction prototype optimization design enhances the effectiveness of BAPFL.

    \item We quantitatively evaluate the performance of BAPFL in PFL based on representative datasets including MNIST \cite{mnist}, FEMNIST \cite{femnist}, and CIFAR-10 \cite{cifar10}. Results show that BAPFL achieves a 35\%-75\% increase in ASR while maintaining ACC. We also integrate BAPFL into different PFL frameworks and heterogeneous settings, and the results highlight its broad adaptability.
\end{itemize}

\section{Related Work}

%\subsection{Prototype-based Federated Learning}
%Federated Learning (FL) enables decentralized model training without sharing raw data, but the data heterogeneity (also known as the non-IID problem) across real-world clients significantly hinders model convergence and accuracy. Existing approaches—such as FedProx \cite{FedProx}, personalized FL \cite{PFedBA, bad-pfl, bapfl}, and cluster-based aggregation \cite{Cluster1, Cluster2}—attempt to mitigate heterogeneity via regularization, personalization, or multiple global models generation. However, these methods rely on gradient-based aggregation, leading to high communication cost and vulnerability to gradient manipulation attacks \cite{tan2021fedproto}.

%To address the challenges posed by data heterogeneity in FL, existing solutions can be categorized into gradient-based and prototype-based approaches. 

To address the challenges posed by data heterogeneity in FL, existing solutions can be categorized into model-based and data-based methods. Model-based methods aim to enhance the final model’s ability to adapt to the diverse data distributions of clients. For instance, FedProx \cite{FedProx} introduces a proximal regularization term to restrict model divergence. Personalized FL approaches \cite{FedALA, PFedBA, bad-pfl, bapfl} enable each client to maintain individualized models to better fit their local data. EAFL \cite{Cluster1} and PrivCrFL \cite{Cluster2} partition clients into groups with similar data distributions and train separate global models per cluster. However, these methods incur high communication overhead. Data-based methods are a more communication-efficient alternative, focusing on learning shared representations across clients. For example, Fed2KD \cite{KD-FL} shares knowledge across clients to boost the model accuracy. GPFL \cite{GPFL} and FedCR \cite{FedCR} extract global and personalized features/representations to enhance model generalization. PFL \cite{tan2021fedproto, FedProc, tan2022federated, FedPA, FedLSA, TAN2025FedPD} leverages class prototypes to align local and global semantics. FPL \cite{FPL} and FedPLVM \cite{FedPLVM} further build hierarchical and unbiased prototypes for better learning performance. In this paper, we focus on PFL and explore its security threats.

Backdoor attacks have proven effective in vanilla FL, and are typically categorized into data poisoning attacks \cite{SADBA2025} and model poisoning attacks \cite{MR, DBA, FCBA}. Data poisoning attacks inject trigger-embedded samples into local datasets to poison local models. In contrast, model poisoning attacks directly manipulate model updates for stronger attack effectiveness. Representative methods include model replacement (MR) \cite{MR}, which scales malicious updates to pollute the aggregated model but suffers from dilution by subsequent benign updates. To enhance the persistence of the attack, distributed backdoor attack (DBA) \cite{DBA} distributes trigger fragments across clients. Full combination backdoor attack (FCBA) \cite{FCBA} further creates diverse trigger variants to increase the ASR. Additionally, PFedBA \cite{PFedBA} and 3DFed \cite{3DFed} incorporate anomaly-aware loss functions to improve attack stealthiness. However, the effectiveness of these backdoor attacks in PFL remains underexplored. In this paper, we fill this gap and propose an effective backdoor attack specifically designed for PFL.

\section{Modeling and Analysis} \label{background}
\subsection{FL Versus PFL}
Consider a FL system with $C$ clients (denoted as \( \mathcal{C} = \{1, \ldots, c, \ldots, C\} \)) and a central server \cite{FL-loss}. Each client \( c \) holds a private dataset \( \mathcal{D}_c = \{(x_c^{i}, y_c^{i})\}_{i=1}^{|\mathcal{D}_c|} \). The local training objective is:
\begin{equation}
\arg\min_\theta\ \mathcal{L}_S = \frac{1}{|\mathcal{D}_c|} \sum\nolimits_{i=1}^{|\mathcal{D}_c|} \ell(f_\theta(x_c^{i}), y_c^{i}),
\end{equation}
where \( \ell(\cdot, \cdot) \) denotes the loss of supervised learning, and $f_\theta$ is the model parameterized by \( \theta \). In each round, clients send model updates to the server for aggregation. However, under heterogeneous data, the aggregated model may perform poorly on some clients. 

PFL \cite{tan2021fedproto} mitigates this issue by exchanging local prototypes, i.e., mean feature vectors, instead of model updates to enhance model generalization. As illustrated in Figure~\ref{fig:pfl}, each client shares a common feature extractor \( \phi(\cdot) \), and computes the local class prototype for class \( k \) as:
\begin{equation}
P_c^{(k)} = \frac{1}{|\mathcal{D}_c^{(k)}|} \sum\nolimits_{(x_c^{i}, y_c^{i}) \in \mathcal{D}_c^{(k)}} \phi(x_c^{i}), 
\end{equation}
where $\mathcal{D}_c^{(k)} = \{(x_c^{i}, y_c^{i}) \in \mathcal{D}_c \mid y_c^{i} = k\}$. Then, the server aggregates local prototypes via:
\begin{equation}
\bar{P}^{(k)} = \sum\nolimits_{c=1}^{C} \frac{|\mathcal{D}_c^{(k)}|}{\sum_{c^{\prime}=1}^{C} |\mathcal{D}_{c^{\prime}}^{(k)}|} P_c^{(k)}.
\end{equation}
Subsequently, client $c$ optimizes its local model using its private data and the global prototypes $\bar{P} = \{\bar{P}^{(k)}\}_{k=1,2,...}$ by minimizing a combined loss $\mathcal{L}$, which includes the supervised loss $\mathcal{L}_S$ and a prototype regularization term $\mathcal{L}_P$, i.e.,
\begin{equation}
\begin{aligned}
\mathcal{L} &= \mathcal{L}_S+\lambda \cdot \mathcal{L}_P\\
              &= \frac{1}{|\mathcal{D}_c|} \sum_{i=1}^{|\mathcal{D}_c|} [\ell(f_\theta(x_c^{i}), y_c^{i}) + \lambda \cdot \| \phi(x_c^i) - \bar{P}^{(y_c^i)} \|_2], \label{eq:pfl_loss}
\end{aligned}
\end{equation}
where $\lambda$ is the coefficient that controls the trade-off between $\mathcal{L}_S$ and $\mathcal{L}_P$.

\subsection{Threat Model}
\subsubsection{Adversary’s Goal} 
Similar to previous backdoor attacks \cite{DBA,SADBA2025}, we consider an adversary that can control multiple compromised clients to upload poisoned prototypes after local training. Its goal is to contaminate benign clients’ models such that they misclassify trigger-embedded samples as the target label, while maintaining high test accuracy on clean samples. The adversary further aims for the backdoor to be stealthy and persistent, avoiding detection and removal throughout training.

\subsubsection{Adversary’s Knowledge and Capability} 
The adversary fully controls the compromised clients, along with their data, training process, and the received global prototypes. However, the adversary cannot control the server and the benign clients. That is, the adversary cannot modify the aggregation rules or interfere with the training process of benign clients. Additionally, we assume that the adversary may obtain knowledge of the local label spaces of benign clients following previous works \cite{lfa2022, SADBA2025}.

\begin{figure}[t]
  \centering
  \subfigure[]{
    \includegraphics[width=0.47\linewidth]{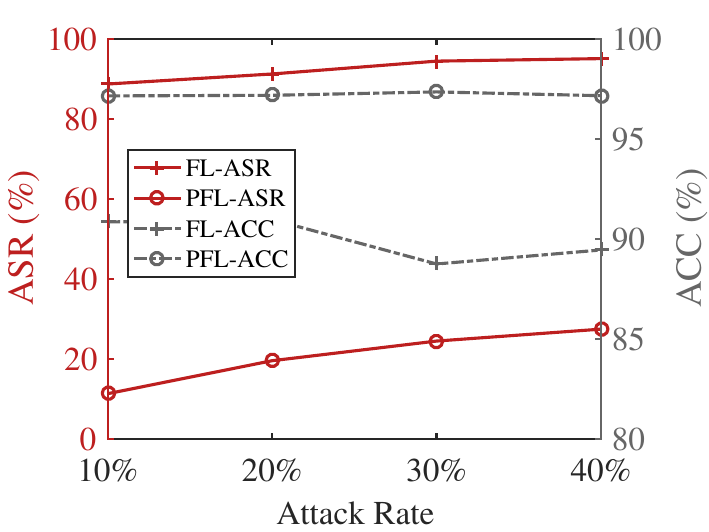}
  }
  \subfigure[]{
    \includegraphics[width=0.47\linewidth]{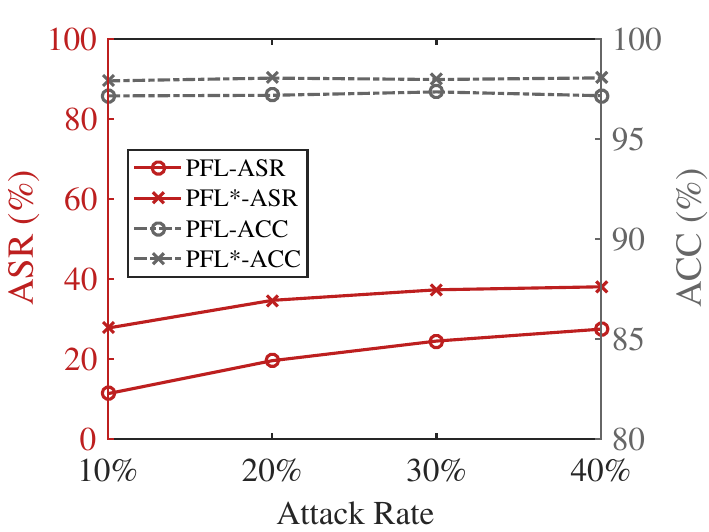}
  }
  \caption{The ASR and ACC of FL, PFL and PFL$^*$ under backdoor attacks based on MNIST.}\label{fig:fl-fedproto}
\end{figure}

\begin{figure}[t]
  \centering
  \subfigure[Round 80]{
    \includegraphics[width=0.4\linewidth]{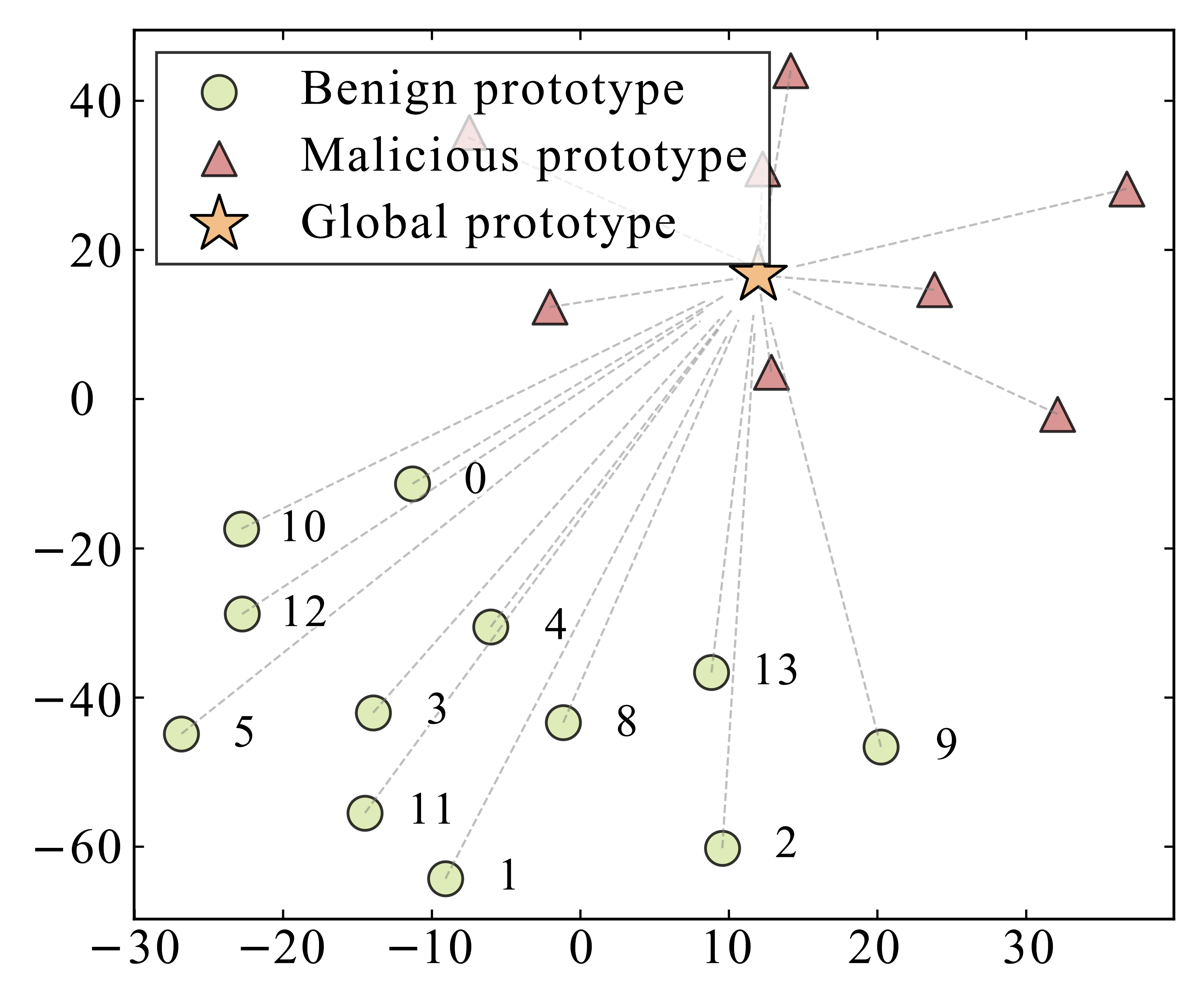} %Round80_poison_label0.png
  }
  \hspace{0.05\linewidth}
  \subfigure[Round 160]{
    \includegraphics[width=0.4\linewidth]{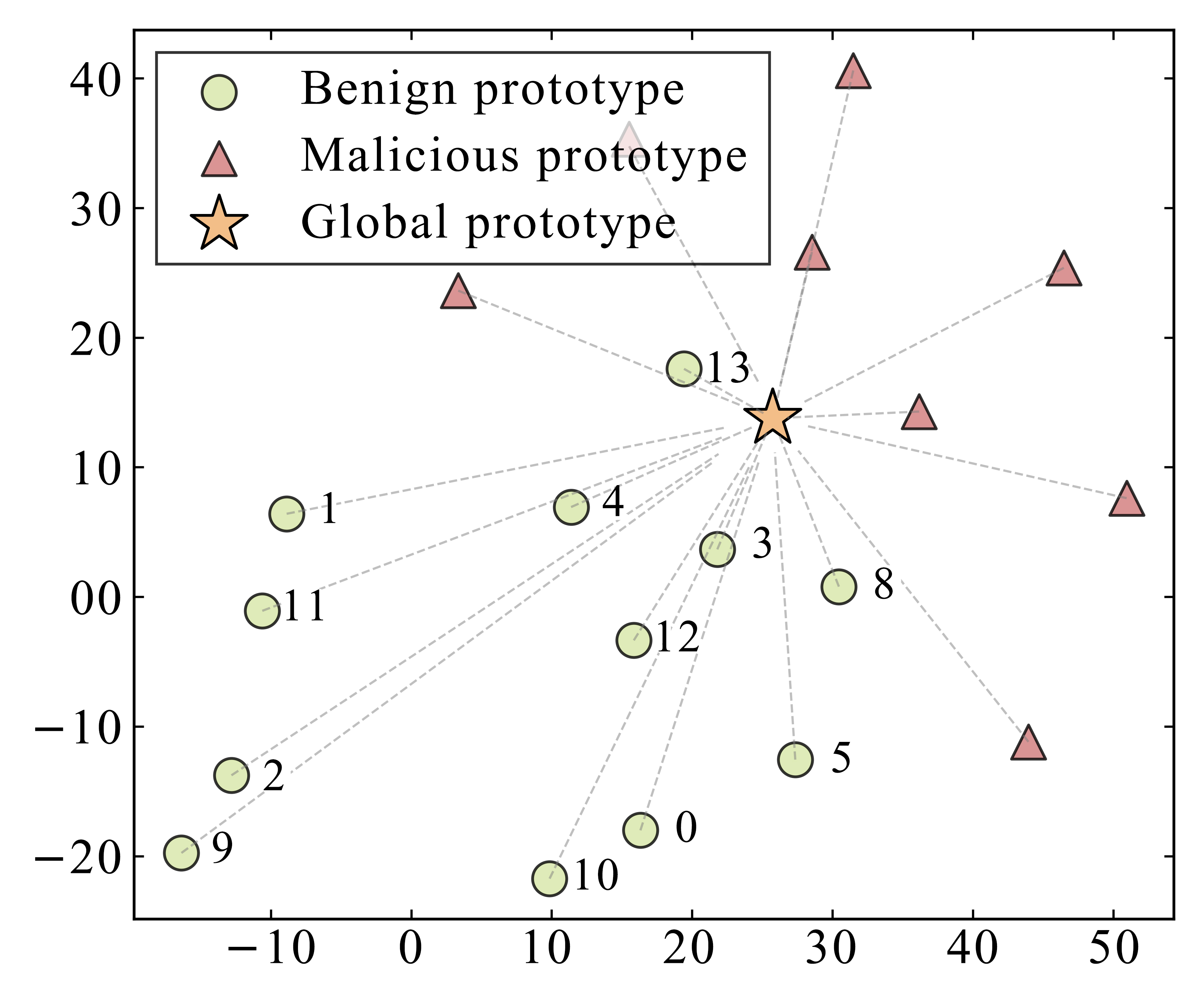} %Round140_poison_label0.png
  }
  \caption{t-SNE visualization of benign and poisoned prototypes at different training rounds. The benign prototypes gradually aligns with the global prototype.}
  \label{fig:benign-global}
\end{figure}

\begin{figure*}[t]
	\centering
	\includegraphics[width=0.75\linewidth]{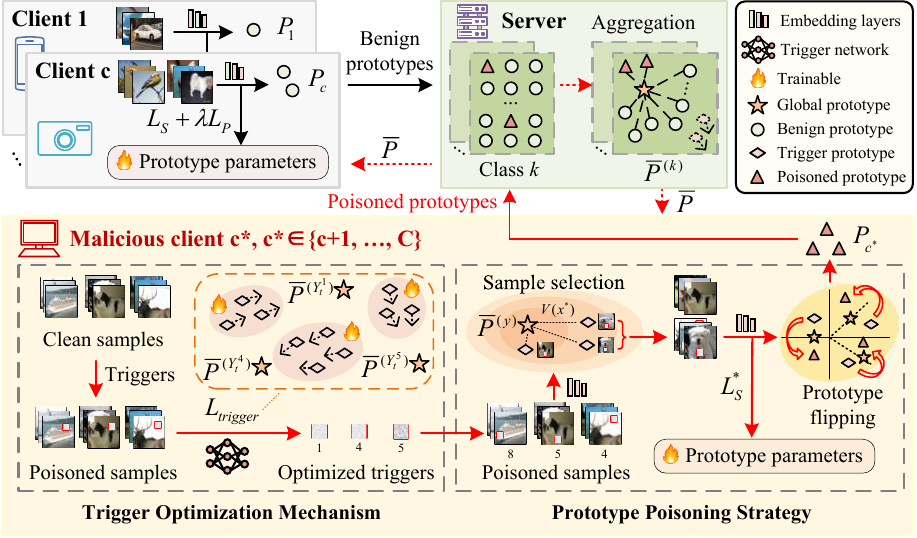}
	\caption{Overview of BAPFL in PFL. }
    \label{fig:bapfl}
\end{figure*}% and Malicious clients $c^*$ executes TOM to optimizes the triggers for the target label set $Y_t$ and performs PPS to generates the poisoned prototypes $P_{c^*}$ to pollute the global prototypes.
%BAPFL consists of two key components: 1) \textit{Prototype Poisoning Strategy}, which generates poisoned prototypes to manipulate the global prototypes away from the trigger prototypes, thereby mislead the benign clients’ local prototypes, and 2) \textit{Trigger Optimization Mechanism}, which optimizes class-specific triggers by aligning their prototypes with the global prototypes of target classes. 
%We consider a standard federated backdoor attack, where each compromised client $c^*$ poisons other benign models by inserting a backdoor task into its local model training process. During training, the malicious client $c^*$ optimizes the following objective:

\subsection{Challenges of Backdoor Attacks in PFL}

We consider a standard federated backdoor attack, where each compromised client $c^*$ poisons other benign models by inserting a backdoor task into its local model training. During training, client $c^*$ minimizes the following loss:
\begin{equation}
\begin{aligned}
\mathcal{L}_S^* =\ & (1 - \alpha) \cdot \frac{1}{|\mathcal{D}_{c^*}|} \sum\nolimits_{i=1}^{|\mathcal{D}_{c^*}|} \ell(f_\theta(x_{c^*}^i), y_{c^*}^i) \\
& + \alpha \cdot \frac{1}{m_{c^*}} \sum\nolimits_{j=1}^{m_{c^*}} \ell(f_\theta(T(x_{c^*}^j)), y_t),
\end{aligned} \label{eq:ba}
\end{equation}
where $T(\cdot)$ is the trigger function that injects a trigger to the training samples and assigns them a target label $y_t$, $m_{c^*}$ is the number of poisoned samples, and $\alpha$ is the poisoning ratio that controls the importance of the backdoor task relative to the main task. In PFL, however, we observe this standard backdoor attack consistently yields low ASR. We identify that this failure stems from  two key factors:

%\subsubsection{The limited influence of poisoned prototypes} In FL, the attacker can effectively embed backdoor effect by poisoning all benign model parameters. In PFL, however, the attacker can only affect the embedding layer of benign models via $\mathcal{L}_P$. Its attack effectiveness are obstructed by the unaffected decision layer of the benign model. To validate this, we assess the performance of the backdoor attack against the standard PFL and FL frameworks. The results are shown in Figure \ref{fig:fl-fedproto}(a). We find that the ASR in FL setting remains above 80\%, while the ASR in PFL reaches only around 10\%-20\%. This indicates that global prototypes exert limited influence on the decision-making of benign clients' model, thereby obstructing the backdoor propagation path.

%\subsubsection{Data heterogeneity of clients} In PFL, some clients do not contain the training samples that belong to the target label $y_t$. Thus their models lack classifier parameters for $y_t$, inherently avoiding the mapping from trigger to $y_t$. We confirm this via an ablation study, in which we inject the training samples of the $y_t$-th class into all clients under PFL (denoted as "PFL$^*$") and compare it with the original PFL. As shown in Figure~\ref{fig:fl-fedproto}(b), the attacker in PFL$^*$ achieves higher ASR across all attack rates. 
\begin{itemize}
    \item \textbf{The limited influence of poisoned prototypes}. In FL, the attacker can effectively embed backdoor effect by poisoning all benign model parameters. In PFL, however, the attacker can only affect the embedding layer of benign models via $\mathcal{L}_P$. Its attack effectiveness is obstructed by the unaffected decision layer of the benign model. To validate this, we assess the performance of the backdoor attack against the standard PFL and FL frameworks. The results are shown in Figure \ref{fig:fl-fedproto}(a). We find that the ASR in FL setting remains above 80\%, while the ASR in PFL reaches only around 10\%-20\%. This indicates that global prototypes exert limited influence on the decision-making of benign clients' models, thereby obstructing the backdoor propagation path.

    \item \textbf{Data heterogeneity of clients}. In PFL, some clients do not contain the training samples that belong to the target label $y_t$. Thus their models lack classifier parameters for $y_t$, inherently avoiding the mapping from trigger to $y_t$. We confirm this via an ablation study, in which we inject the training samples of $y_t$ into all clients under PFL (denoted as “PFL$^*$”) and compare it with the original PFL. As shown in Figure~\ref{fig:fl-fedproto}(b), the attacker in PFL$^*$ achieves higher ASR across all attack rates. 
\end{itemize}
%In PFL framework (e.g., FedProto~\cite{tan2021fedproto}), benign clients treat global prototypes as regularization signals rather than direct parameters. Their local training is still dominated by private data, which reduces the probability of learning the malicious prototype-to-label mapping injected by the attacker. To validate this, we compare two settings: one where clients perform standard FedProto training, and another where clients perform standard FL training. The results are shown in Figure \ref{fig:fl-fedproto}(a). We find that the ASR in the latter setting remains above 80\%, while FedProto with prototype regularization reaches only around 10-20\%. This indicates that global prototypes exert limited influence on the benign clients' model updates, thereby weakening the backdoor propagation path.

%Limited dependence on global prototypes during local training
%Label heterogeneity over clients

%1) The limited influence of poisoned prototypes. Even if the global prototypes are contaminated by poisoned prototypes, they only affect the embedding layer of benign models. Since the benign models are trained with clean dataset, their decision layer still learns to classify perturbed local prototypes into correct labels.

%The above observations reveal that both the training paradigm of PFL and the data heterogeneity among clients collectively reduce the effectiveness of conventional backdoor attacks. 
%We consider a simpler target for a backdoor attack

The above challenges motivate us to rethink backdoor attack strategies in PFL. We first examine the distribution changes of global and benign prototypes at the 80-th and 160-th training rounds in PFL under backdoor attacks, and the results are shown in Figure \ref{fig:benign-global}. We observe that, as the number of training rounds increases, benign prototypes gradually converge toward the manipulated global prototype. This motivates us to develop a novel attack strategy: \textit{\textbf{by manipulating the global prototype away from the trigger prototype, the attacker may indirectly push benign prototypes away from the trigger prototypes, thereby increasing the probability of misclassifying trigger-embedded samples}}. To further classify the trigger-embedded samples into the target label, the trigger prototypes can be optimized toward the global prototype of the target label.

\section{Proposed BAPFL: Backdoor Attack Against Prototype-based Federated Learning} \label{sec:bapfl}
\subsection{Overview}
Based on the challenges and potential attack strategy outlined in the previous section, we propose a novel backdoor attack method BAPFL, which exploits the dual-direction prototype optimization mechanism to indirectly propagate backdoor behavior across diverse PFL frameworks~\cite{tan2021fedproto,tan2022federated,TAN2025FedPD}. As illustrated in Figure~\ref{fig:bapfl} and Algorithm \ref{alg:bapfl}, BAPFL integrates two components: \textit{prototype poisoning strategy} and \textit{trigger optimization mechanism}. Each malicious client \( c^* \) executes PPS and TOM to generate poisoned prototypes \( P_{c^*} \) and optimize label-specific triggers, respectively. Specifically, the poisoned prototypes of PPS deliberately bias the global prototypes away from the trigger prototypes. This manipulation indirectly influences the prototype learning of benign clients, pushing their benign prototypes to diverge from the trigger prototypes. As this discrepancy increases, trigger-embedded samples are more likely to be misclassified by benign models. Meanwhile, TOM expands the target label space and optimizes triggers for each target label. This increases the probability that the target labels overlap with the local label space of benign clients, thereby enabling the benign models to inadvertently activate the \textit{trigger-target label} mapping. Further theoretical analysis of BAPFL is provided in Appendix~A.
\subsection{Prototype Poisoning Strategy (PPS)}
PPS includes two steps: sample selection and prototype flipping. The former identifies the most valuable trigger-embedded samples. The prototypes of these selected samples serve as the basis for the latter, which constructs the accurate poisoned prototypes in a deliberately opposite direction, thereby manipulating the global prototypes away from the trigger prototypes.

\subsubsection{Sample Selection Strategy}
Malicious clients first compute the attack value of each trigger-embedded sample $x^*$ by computing the Euclidean distance between its prototype $\phi(x^*)$ and the global prototype $\bar{P}^{(y)}$, where $y$ is the ground-truth label of the clean sample $x$ related to $x^*$. That is, %i.e., % of the original class $y$. 
    \begin{equation}
    V(x^*) = \| \phi(x^*) - \bar{P}^{(y)} \|_2, \label{eq:sample-select}
    \end{equation}
    where a larger distance implies higher attack value. The top-$K$ samples with the highest attack values are selected for training local model and constructing poisoned prototypes.

\subsubsection{Prototype Flipping Strategy} 
To mislead the global prototype, malicious clients construct poisoned prototypes and upload them to the server. Specifically, malicious client $c^*$ first computes the class-wise trigger prototype $P_{tr}^{(k)}$ from the selected samples. Then, $c^*$ computes the projection of $P_{tr}^{(k)}$ onto the corresponding global prototype $\bar{P}^{(k)}$. Finally, $c^*$ constructs the poisoned prototype $P_{c^*}^{(k)}$ by performing a symmetrical flip of $P_{tr}^{(k)}$ with respect to this projection, i.e.,
    \begin{equation}
        P_{c^*}^{(k)} = 2 \cdot P_{proj} - P_{tr}^{(k)}, \label{eq:proto-flip}
    \end{equation}
    where $P_{proj} = \frac{\bar{P}^{(k)} \cdot P_{tr}^{(k)}}{\bar{P}^{(k)} \cdot \bar{P}^{(k)}} \cdot \bar{P}^{(k)}$ denotes the projection of $P_{tr}^{(k)}$ onto $\bar{P}^{(k)}$. Compared with other flipping strategies such as origin-based or global prototype-based symmetry, our proposed prototype flipping strategy achieves finer control over both the direction and the norm of poisoned prototypes (see Appendix B). This ensures both effective and stealthy manipulation of benign clients’ prototype learning. 

\begin{algorithm}[tb]
\caption{PFL process with BAPFL Attack}
\label{alg:bapfl}
\begin{algorithmic}[1] %[1] enables line numbers
\STATE \textbf{Server Executes:} 
\STATE Initialize global prototypes $\bar{P} = \{\bar{P}^{(k)}\}_{k=1,2,...}$\;
\WHILE{\textit{the current training round $\leq$ the final round}}
\STATE Broadcast $\bar{P}$ to clients for local training
\STATE Aggregate the local prototypes of clients to update $\bar{P}$\;
\ENDWHILE
\STATE \textbf{Client Executes:} 
\IF {\textit{this client is compromised}}
\STATE /*\textit{Execute TOM}*/
\STATE Triggers $\gets$ Download the trigger network from the adversary and train it with $\bar{P}$ based on Equation \eqref{eq:triggerloss}
\STATE /*\textit{Execute PPS}*/
\STATE Select the top-$K$ samples with the highest attack value based on Equation \eqref{eq:sample-select}
\STATE Train $f_\theta$ according to Equation \eqref{eq:ba}
\STATE Poisoned prototypes $P_c \gets$ Flip trigger prototypes
\ELSE
\STATE Train $P_c$ and $f_\theta$ with $\bar{P}$ according to Equation \eqref{eq:pfl_loss}
\ENDIF
\STATE \textbf{return} the new prototypes $P_c$ to the server
\end{algorithmic}
\end{algorithm}

\subsection{Trigger Optimization Mechanism (TOM)}
TOM includes two steps: trigger optimization and trigger training. The former designates a specific trigger to each target label, while the latter trains these triggers to achieve optimal effectiveness and stealthiness.
%To enhance the attack's adaptability and effectiveness across diverse clients, we introduce a Trigger Optimization Mechanism (TOM) that constructs learnable and class-specific triggers. These triggers are jointly optimized to maximize the backdoor effect under the constraints of sparsity and stealthiness. The TOM mechanism consists of the following two modules:

\subsubsection{Trigger Optimization Strategy} 
To enhance attack effectiveness across benign clients with heterogeneous data, we expand the attack's target label space from a single label $y_t$ to a label set $Y_t$ that encompasses all local labels of benign clients. For each target label $y_t \in Y_t$, a specific trigger $(\delta_{y_t}, M_{y_t})$ is learned. This design enables BAPFL to perform personalized backdoor attacks on benign clients. Specifically, if the local label space of benign client $c$ contains the target label $y_t$, BAPFL can activate backdoor behaviors of $c$’s local model, enabling it to classify the samples embedded with the trigger $(\delta_{y_t}, M_{y_t})$ as $y_t$.
    
    % regardless of their local label distribution, since the malicious prototypes have poisoned all class prototypes in the global prototype. Specifically, during local training, malicious clients inject backdoor samples with the corresponding class-specific triggers and upload poisoned prototypes to the server. Because $Y_t$ includes all class labels, there always exists at least one target label that overlaps with the label space of each benign client. As a result, the corresponding benign prototypes are inevitably influenced by the poisoned global prototypes. This ensures that benign clients are likely to learn the trigger-to-label mapping during their local training.

\subsubsection{Trigger Training Strategy}
For each target label $y_t$, we learn a dedicated trigger pattern $\delta_{y_t}$ and a corresponding mask $M_{y_t}$, forming a trigger function $T_{y_t}(x) = (1 - M_{y_t}) \odot x + M_{y_t} \odot \delta_{y_t}$. The optimization objective of each trigger aims to simultaneously: 1) minimize classification loss of $T_{y_t}(x)$ to $y_t$, 2) align the prototype of $T_{y_t}(x)$ with the global prototype $\bar{P}^{(y_t)}$, and 3) ensure visual imperceptibility of the trigger. The corresponding loss function is formulated as:
\begin{equation}
\begin{aligned}
\mathcal{L}_{trigger} =\ & \underbrace{\mathcal{L}_{\text{S}}(f_\theta(T_{y_t}(x)), y_t)}_{\text{target classification loss}} 
+ \lambda_1 \cdot \underbrace{\| \phi(T_{y_t}(x)) - \bar{P}^{(y_t)} \|_2}_{\text{prototype alignment}} \\
& + \lambda_2 \cdot \underbrace{\| M_{y_t} \|_1 + \lambda_3 \cdot \| \delta_{y_t} \|_2}_{\text{stealthiness loss}},
\end{aligned} \label{eq:triggerloss}
\end{equation}
where $\lambda_1$, $\lambda_2$, and $\lambda_3$ are hyperparameters controlling the trade-off between effectiveness and stealthiness.

%\end{itemize}

%Such alignment significantly enhances the stealth and effectiveness of the attack across heterogeneous and disjoint client populations.

    %Specifically, During local training, malicious clients inject backdoor samples with the corresponding class-specific triggers and upload poisoned prototypes accordingly.

\section{Performance Evaluation} %Experiments
%This section provides a detailed description of the implementation and evaluation of our proposed BAPFL. We first present the experimental setup. Then, we compare the backdoor effectiveness of BAPFL with that of existing mainstream backdoor attack methods across three representative classification tasks. Experimental results demonstrate that BAPFL not only achieves significantly higher ASR but also exhibits strong robustness and adaptability across different PFL scenarios.

\begin{table}[t]
\centering
\small
\setlength{\tabcolsep}{1mm}
\resizebox{\linewidth}{!}{  % 缩放开始
\begin{tabular}{lcccc}
\toprule
\multirow{2}{*}{Dataset} & \multirow{2}{*}{Labels} & \multirow{2}{*}{Image size} & Training & \multirow{2}{*}{Model} \\
     &  &  & /Test images &  \\
\midrule
MNIST & 10 & 1*28*28 & 60k/10k & 2Conv + 2Fc \\
FEMNIST & 10 & 1*128*128 & 22k/3k & 2Conv + 2Fc \\
CIFAR-10 & 10 & 3*32*32 & 50k/10k & 2Conv + Pool + 3Fc \\
%OFFICE-10 & 10 & 3*64*64 & 2800/700 & ResNet18 \\
\bottomrule
\end{tabular}
}  % 缩放结束
\caption{Dataset and model architecture} \label{tab:dataset}
\end{table}

\subsection{Experiment Setup}
\subsubsection{Datasets and Models}
Our experiments are conducted on three datasets: MNIST \cite{mnist}, FEMNIST \cite{femnist}, and CIFAR-10 \cite{cifar10}, which are benchmark datasets for image classification. The dataset and model details are provided in Table \ref{tab:dataset}. 

\subsubsection{Training Setting}
In our experiments, we choose FedProto \cite{tan2021fedproto} as the basic PFL framework. The number of clients is set to 20 and the total training rounds are set to 200. In each training round, each client performs 1 local epoch with a local batch size of 4, and the learning rate is set to 0.01. We assume that all clients perform learning tasks with heterogeneous statistical distributions. Specifically, each client is assigned a $p$-way $q$-shot classification task, where $p$ and $q$ denote the number of local classes and the number of samples per class, respectively. We also introduce a standard deviation $std$ to $p$ and $q$ to achieve enhanced heterogeneity for different clients. By default, we set $p=5$, $q=100$, and $std=2$. Additionally, we set $\alpha = 0.75, \lambda = 1, \lambda_1 = 0.1, \lambda_2 = 0.01$ and $\lambda_3 = 0.001$. All experiments are repeated 10 times, and the average results are reported.
%For CIFAR10 dataset, we set $p=4$, $q=100$, and $std=1$. We set $p=5$, $q=100$, and $std=2$ as the default configuration.   with different random seeds

%For all experiments, We also introduce standard deviation $std$ to $p$ and $q$ to achieve enhanced heterogeneity in label space and data quantity for different clients.

\begin{figure*}[htbp]
  \centering
  \subfigure[Different number of local classes]{
    \includegraphics[width=0.25\linewidth]{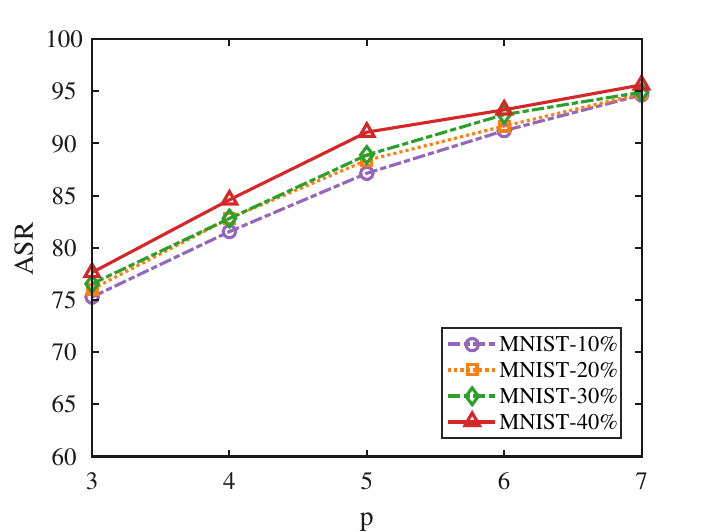}
  }
  \hspace{0.04\linewidth}
  \subfigure[Different number of samples]{
    \includegraphics[width=0.25\linewidth]{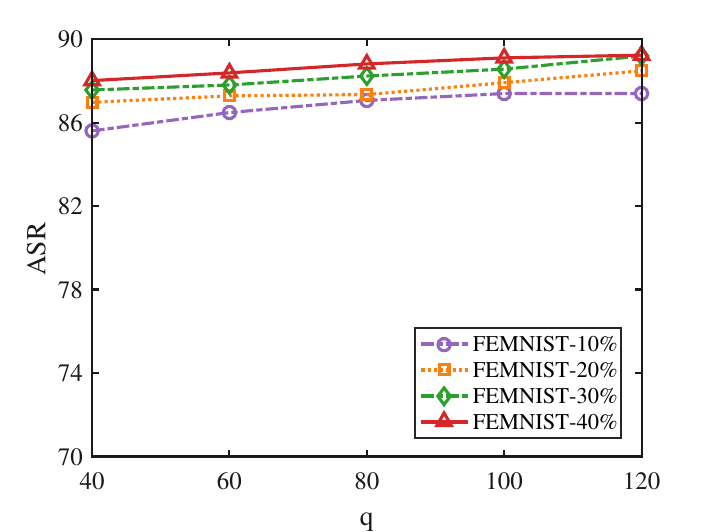}
  }
  \hspace{0.04\linewidth}
  \subfigure[Different standard deviation]{
    \includegraphics[width=0.25\linewidth]{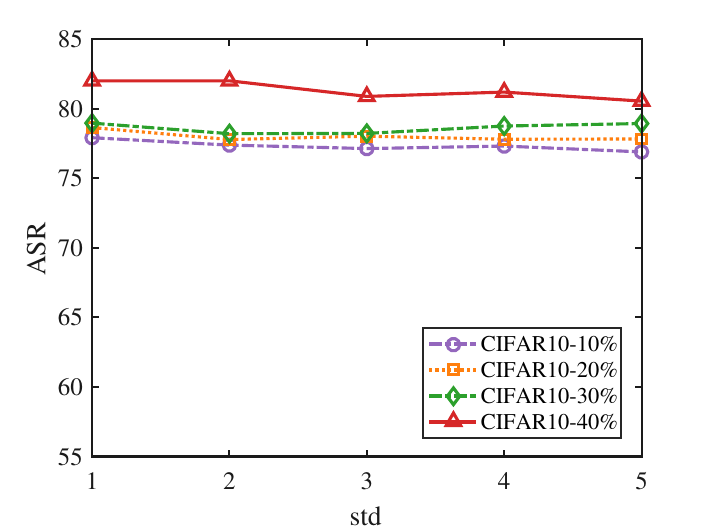}
  }
  \caption{The ASR of BAPFL under varying degrees of data heterogeneity.}
  \label{fig:data_heterogeneity}
\end{figure*}

\subsubsection{Baselines}
To evaluate the effectiveness of BAPFL, we compare it with two representative backdoor attack baselines: MR and DBA. We also assess the adaptability of BAPFL across other PFL frameworks, including FedPD \cite{TAN2025FedPD} and FedPCL \cite{tan2022federated}. 

\subsubsection{Attack Setup}
We simulate a backdoor attack scenario where the attacker controls 10\%, 20\%, 30\%, or 40\% of the clients. For each ground-truth label $y$, the target label is set to $9 - y$. The compromised clients train triggers for their target labels over $50$ local rounds and embed them into their local data to construct poisoned prototypes. During the training process of compromised clients, we set the trigger and the local model to be trained alternately. 
%These poisoned updates are then uploaded to the server during the federated training process.

\subsubsection{Evaluation Metrics} 
We report the average main task accuracy (ACC, \%) over clean samples and average attack success rate (ASR, \%) over trigger-embedded samples for all benign clients’ models on their test datasets. 
%We report the mean main task accuracy (Acc, \%) over clean samples and mean attack success rate (ASR, \%) over trigger-embedded samples for all benign clients’ benign models on their test sets.

% To comprehensively evaluate the performance of our proposed attack, we adopt two key metrics: 

% \begin{itemize}
%     \item \textbf{Main task accuracy (ACC)}. ACC measures the model's prediction accuracy on clean, unmodified test data. It reflects the overall classification performance and serves as an indicator of whether the model maintains normal utility after the backdoor injection.
%     \item \textbf{Attack success rate (ASR)}. ASR quantifies the effectiveness of the backdoor attack. It is defined as the proportion of trigger-embedded samples that are misclassified into the target class as intended by the attacker. A higher ASR indicates a more successful and stealthy backdoor attack.
% \end{itemize}

%Together, these two metrics allow us to evaluate both the stealthiness and the impact of the attack: an ideal backdoor attack achieves a high ASR while maintaining minimal degradation in test accuracy.

\begin{table}[t]
\centering
\setlength{\tabcolsep}{1mm}
\small
\resizebox{\linewidth}{!}{  % 缩放开始
\begin{tabular}{c|cc|cc|cc|cc}
\toprule
\multirow{2}{*}{\textbf{Method}} & \multicolumn{2}{c|}{ \textbf{AR = 10\%}} & \multicolumn{2}{c|}{\textbf{AR = 20\%}} & \multicolumn{2}{c|}{\textbf{AR = 30\%}} & \multicolumn{2}{c}{\textbf{AR = 40\%}} \\
\cmidrule{2-9}
   & ACC & ASR & ACC & ASR & ACC & ASR & ACC & ASR \\
\toprule
    \multicolumn{9}{c}{\textbf{MNIST}}\\
  MR   & 96.79 & 13.29 & 97.39 & 24.54 & 97.47 & 40.27 & 97.89 & 50.99 \\
  DBA  & 97.60 & 38.52 & 97.51 & 42.44 & 96.04 & 49.94 & 98.08 & 56.40 \\
  BAPFL & 97.96 & \textbf{87.14} & 97.85 & \textbf{88.38} & 96.89 & \textbf{88.89} & 96.90 & \textbf{91.08} \\
\midrule
 \multicolumn{9}{c}{\textbf{FEMNIST}}\\
  MR   & 90.97 & 13.68 & 91.31 & 34.18 & 90.17 & 18.73 & 88.31 & 28.52 \\
  DBA  & 89.80 & 11.71 & 91.21 & 17.96 & 89.83 & 21.10 & 89.41 & 41.67 \\
  BAPFL & 91.94 & \textbf{87.39} & 91.29 & \textbf{88.48} & 90.55 & \textbf{89.19} & 89.18 & \textbf{89.23} \\
\midrule
   \multicolumn{9}{c}{\textbf{CIFAR-10}}\\
  MR   & 66.10 & 11.32 & 63.93 & 13.08 & 60.75 & 13.36 & 66.31 & 13.81 \\
  DBA  & 65.97 & 10.25 & 60.31 & 10.63 & 65.71 & 13.48 & 66.44 & 13.59 \\
  BAPFL & 62.38 & \textbf{77.38} & 61.47 & \textbf{77.78} & 60.93 & \textbf{78.20} & 60.83 & \textbf{82.00} \\
\bottomrule
\end{tabular}
}  % 缩放结束
\caption{ACC and ASR of BAPFL and baselines in PFL.}
\label{tab:main_results}
\end{table}

%under different attack rates on MNIST, FEMNIST, and CIFAR-10
\subsection{Main Results}
\subsubsection{Comparisons Between BAPFL and Baselines}
%Based on the classical PFL framework FedProto \cite{tan2021fedproto}, we compare the proposed BAPFL with two representative backdoor attacks, i.e., MR and DBA, under varying attack rates (AR) on MNIST, FEMNIST, and CIFAR-10 datasets. As shown in Table~\ref{tab:main_results}, BAPFL consistently achieves the highest ASR across all settings, while maintaining comparable or even higher ACC. Notably, BAPFL improves ASR by 35\%-75\% over BA and DBA, demonstrating its superior effectiveness and stealthiness in PFL scenarios.

In the PFL framework FedProto, we compare the performance of BAPFL with baselines, i.e., MR and DBA, under varying attack rates (AR) on MNIST, FEMNIST, and CIFAR-10. As shown in Table~\ref{tab:main_results}, BAPFL consistently achieves the highest ASR across all settings, while maintaining comparable or even higher ACC. Notably, BAPFL improves ASR by 35\%–75\% over MR and DBA, demonstrating its superior effectiveness and stealthiness in PFL.

%All experiments are repeated 10 times with different random seeds, and the average results are shown in Table~\ref{tab:main_results}. We observe that BAPFL
%0%的情况去掉
% \begin{table}[t]
% \centering
% \setlength{\tabcolsep}{1mm}
% \caption{BAPFL performance under different PFL frameworks.}
% \label{tab:fedpcl_fedpd}
% \small
% \resizebox{\linewidth}{!}{  % 缩放开始
% \begin{tabular}{l|cc|cc|cc|cc|cc}
% \toprule
% \multirow{2}{*}{\textbf{Method}} & \multicolumn{2}{c|}{\textbf{AR = 0\%}} & \multicolumn{2}{c|}{\textbf{AR = 10\%}} & \multicolumn{2}{c|}{\textbf{AR = 20\%}} & \multicolumn{2}{c|}{\textbf{AR = 30\%}} & \multicolumn{2}{c}{\textbf{AR = 40\%}} \\
% \cmidrule{2-11}
%    & ACC & ASR & ACC & ASR & ACC & ASR & ACC & ASR & ACC & ASR \\
% \midrule
% \multirow{1}{*}{FedPCL}
%   & 49.69 & - & 49.11 & 72.91 & 48.64 & 75.89 & 48.61 & 78.78 & 49.81 & 81.82 \\
% \midrule
% \multirow{1}{*}{FedPD} & 97.97 & - & 97.87 & 65.11 & 97.91 & 70.56 & 96.97 & 77.74 & 97.28 & 79.40 \\
% \bottomrule
% \end{tabular}
% }
% \end{table}

\begin{table}[t]
\centering
\setlength{\tabcolsep}{1mm}
\small
\resizebox{\linewidth}{!}{  % 缩放开始
\begin{tabular}{l|cc|cc|cc|cc}
\toprule
\multirow{2}{*}{\textbf{Method}}  & \multicolumn{2}{c|}{\textbf{AR = 10\%}} & \multicolumn{2}{c|}{\textbf{AR = 20\%}} & \multicolumn{2}{c|}{\textbf{AR = 30\%}} & \multicolumn{2}{c}{\textbf{AR = 40\%}} \\
\cmidrule{2-9}
   &  ACC & ASR & ACC & ASR & ACC & ASR & ACC & ASR \\
\toprule
\multirow{1}{*}{FedPCL}
  &  49.11 & 72.91 & 48.64 & 75.89 & 48.61 & 78.78 & 49.81 & 81.82 \\
\midrule
\multirow{1}{*}{FedPD} & 97.87 & 65.11 & 97.91 & 70.56 & 96.97 & 77.74 & 97.28 & 79.40 \\
\bottomrule
\end{tabular}
}
\caption{BAPFL performance in various PFL frameworks.}
\label{tab:fedpcl_fedpd}
\end{table}

\subsubsection{Integrate BAPFL into Different PFL Frameworks}

To demonstrate the adaptability of BAPFL, we evaluate its effectiveness against two other representative PFL frameworks: FedPCL and FedPD. FedPCL employs a contrastive loss to enhance prototype alignment. FedPD adopts robust aggregation based on cosine similarity and encourages inter-class prototype separation. We apply BAPFL to FedPCL on the OFFICE-10 dataset~\cite{office}, and to FedPD on MNIST, respectively. The results are shown in Table~\ref{tab:fedpcl_fedpd}. We observe that as the attack rate increases, the ASR of BAPFL in FedPCL increases from 72.91\% to 81.82\%, while ACC remains stable. This confirms the vulnerability of FedPCL to our attack. Moreover, although FedPD adopts robust aggregation, BAPFL still achieves 65.11\%-79.4\% ASR, demonstrating its ability to bypass FedPD’s defense.

%with minimal ACC loss
%FedPCL defines its loss as the $\ell_2$ distance between local and global prototypes.

\subsubsection{Data Heterogeneity}

To evaluate the robustness of BAPFL under varying degrees of data heterogeneity, we simulate different data heterogeneity scenarios by adjusting the values of $p$, $q$, and $std$ across clients. Specifically, for MNIST, we fix $q=100$, $std=2$, and vary $p$ from 3 to 7. For FEMNIST, we fix $p=5$, $std=2$, and vary $q$ from 40 to 120. For CIFAR-10, we fix $p=5$, $q=100$, and vary $std$ from 1 to 5. Then, we evaluate the ASR of our BAPFL method under these settings. The experimental results in Figure \ref{fig:data_heterogeneity} show that BAPFL consistently achieves ASR of at least 75\% across all heterogeneous settings, demonstrating its strong robustness.

\subsubsection{The Effect of $\lambda$ for BAPFL}
%To investigate the impact of the prototype loss weight $\lambda$ on the performance of BAPFL, we evaluate the attack success rate (ASR) and accuracy (ACC) on the MNIST dataset under different values of $\lambda \in \{1, 2, 3, 4, 5\}$. Here, $\lambda$ controls the relative importance of the prototype alignment loss for benign clients during local training. 

%To assess the effect of the weight $\lambda$ of $\mathcal{L}_P$ for BAPFL, we evaluate the ASR and ACC of BAPFL in PFL with $\lambda \in \{1, 2, 3, 4, 5\}$. Figure~\ref{fig:protoype_loss} presents the experimental results on the MNIST dataset. We observe that the ASR results remain stable, while $\lambda$ has a relatively noticeable effect on the ACC results. Specifically, the ASR of BAPFL remains consistently high across all settings, with only minor fluctuations across different $\lambda$ values. In contrast, the ACC of BAPFL decreases as $\lambda$ increases under high attack rates. For instance, the ACC value drops from 97\% ($\lambda=1$) to 92\% ($\lambda=5$) when the attack rate is 40\%. 

\begin{figure}[t]
  \centering
  \subfigure[]{
    \includegraphics[width=0.47\linewidth]{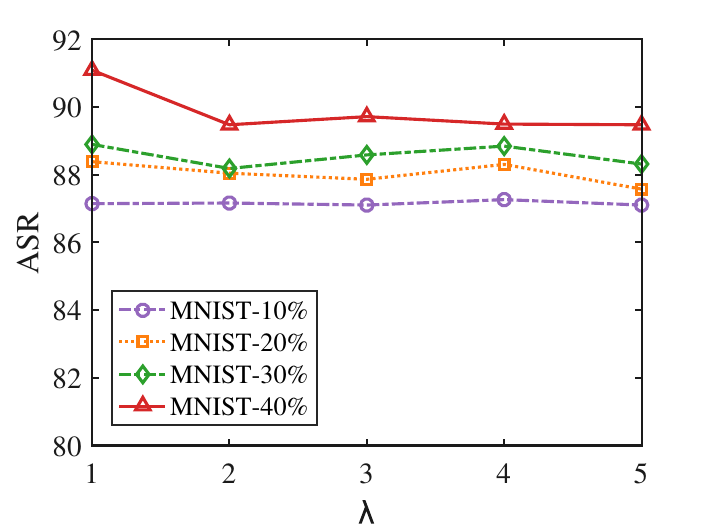}
  }
  \subfigure[]{
    \includegraphics[width=0.47\linewidth]{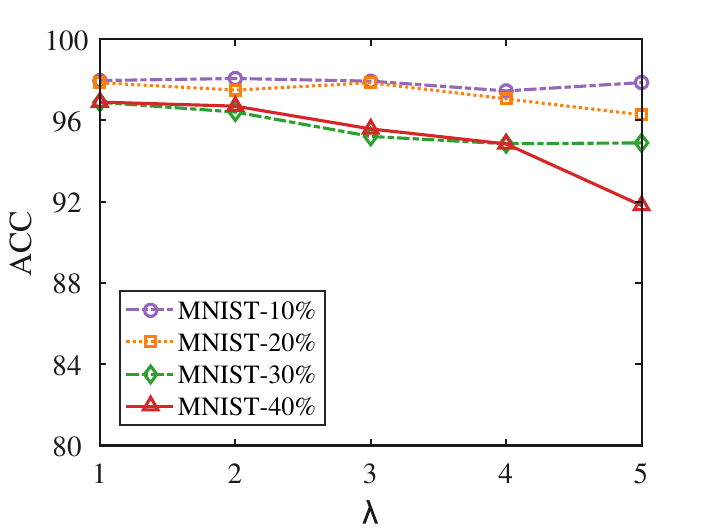}
  }
  \caption{Effect of the weight $\lambda$ on ASR and ACC of BAPFL on the MNIST dataset.}
  \label{fig:protoype_loss}
\end{figure}

To assess the effect of the weight $\lambda$ of $\mathcal{L}_P$ for BAPFL, we evaluate the ASR and ACC of BAPFL in PFL with $\lambda \in \{1, 2, 3, 4, 5\}$. Figure~\ref{fig:protoype_loss} presents the experimental results on the MNIST dataset. We observe that BAPFL achieves consistently high ASR across all settings, with only minor fluctuations across different $\lambda$ values. Moreover, the ACC of BAPFL generally remains stable. However, under high attack rates, the ACC of BAPFL decreases slightly as $\lambda$ increases. For example, when the attack rate is 40\%, the ACC value drops from 97\% ($\lambda=1$) to 92\% ($\lambda=5$). The attacker can mitigate this effect by reducing the attack rate. Overall, BAPFL demonstrates its effectiveness under different $\lambda$ settings.

%This indicates that strengthening the alignment of benign prototypes and polluted global prototypes may harm the accuracy of benign models.
%prototype alignment

%the ASR of BAPFL stays consistently high across all settings. As the attack rate increases, the ASR value increases from 87\% to 91\%, with only minor fluctuations across different $\lambda$ values. In contrast, the ACC of BAPFL drops more significantly with larger $\lambda$. For example, at a 40\% attack rate, the ACC value decreases from 97\% ($\lambda=1$) to 92\% ($\lambda=5$), indicating that excessive emphasis on prototype alignment can harm model accuracy.

% \begin{table*}[t]
% \centering
% \caption{Ablation results of BAPFL on MNIST, FEMNIST, and CIFAR-10 datasets under different attack rates.}
% \label{tab:ablation}
% \small
% \begin{tabular}{ll|cc|cc|cc|cc}
% \toprule
% \multirow{2}{*}{\textbf{Dataset}} & \multirow{2}{*}{\textbf{Method}} & \multicolumn{2}{c|}{\textbf{AR = 10\%}} & \multicolumn{2}{c|}{\textbf{AR = 20\%}} & \multicolumn{2}{c|}{\textbf{AR = 30\%}} & \multicolumn{2}{c}{\textbf{AR = 40\%}} \\
% \cmidrule{3-10}
% & & ACC & ASR & ACC & ASR & ACC & ASR & ACC & ASR \\
% \midrule
% \multirow{3}{*}{MNIST} 
% & Ours (TTS)        & 98.17 & 63.55 & 98.19 & 64.91 & 98.36 & 69.55 & 98.17 & 73.60 \\
% & Ours (TTS+PPS)    & 98.05 & 66.44 & 97.98 & 69.89 & 97.72 & 71.50 & 97.48 & 75.64 \\
% & Ours (TOS+PPS)    & 97.86 & \textbf{82.15} & 97.39 & \textbf{87.57} & 96.18 & \textbf{88.18} & 96.79 & \textbf{89.40} \\
% \midrule
% \multirow{3}{*}{FEMNIST} 
% & Ours (TTS)        & 90.69 & 73.01 & 90.46 & 74.91 & 88.37 & 76.89 & 88.25 & 77.02 \\
% & Ours (TTS+PPS)    & 90.00 & 74.23 & 89.93 & 77.25 & 89.32 & 78.63 & 88.91 & 79.84 \\
% & Ours (TOS+PPS)    & 91.69 & \textbf{81.88} & 91.06 & \textbf{86.38} & 89.93 & \textbf{86.60} & 89.56 & \textbf{87.56} \\
% \midrule
% \multirow{3}{*}{CIFAR-10} 
% & Ours (TTS)        & 63.23 & 42.86 & 66.63 & 45.58 & 65.94 & 45.67 & 65.74 & 45.84 \\
% & Ours (TTS+PPS)    & 65.59 & 45.64 & 67.21 & 47.85 & 61.87 & 49.93 & 65.16 & 50.69 \\
% & Ours (TOS+PPS)    & 63.54 & \textbf{73.04} & 61.45 & \textbf{73.19} & 61.33 & \textbf{73.21} & 64.67 & \textbf{73.52} \\
% \bottomrule
% \end{tabular}
% \end{table*}

\begin{table}[t]
\centering
\setlength{\tabcolsep}{1mm}
\small
\resizebox{\linewidth}{!}{  % 缩放开始
\begin{tabular}{l|cc|cc|cc|cc}
\toprule
 \multirow{2}{*}{\textbf{Method}} & \multicolumn{2}{c|}{\textbf{AR = 10\%}} & \multicolumn{2}{c|}{\textbf{AR = 20\%}} & \multicolumn{2}{c|}{\textbf{AR = 30\%}} & \multicolumn{2}{c}{\textbf{AR = 40\%}} \\
\cmidrule{2-9}
 & ACC & ASR & ACC & ASR & ACC & ASR & ACC & ASR \\
\toprule
\multicolumn{9}{c}{\textbf{MNIST}} \\
 DBA        & 97.60 & 38.52 & 97.51 & 42.44 & 96.04 & 49.94 & 98.08 & 56.40 \\
 DBA+PPS    & 98.05 & 62.15 & 97.98 & 69.43 & 97.72 & 71.50 & 97.48 & 75.64 \\  
 PPS+TOM   & 97.96 & \textbf{87.14} & 97.85 & \textbf{88.38} & 96.89 & \textbf{88.89} & 96.90 & \textbf{91.08} \\
\midrule
\multicolumn{9}{c}{\textbf{FEMNIST}} \\
 DBA        & 89.80 & 11.71 & 91.21 & 17.96 & 89.83 & 21.10 & 89.41 & 41.67 \\
 DBA+PPS    & 89.80 & 60.25 & 90.09 & 69.72 & 89.72 & 72.72 & 88.91 & 73.58 \\ 
 PPS+TOM    & 91.94 & \textbf{87.39} & 91.29 & \textbf{88.48} & 90.55 & \textbf{89.19} & 89.18 & \textbf{89.23} \\
\midrule
\multicolumn{9}{c}{\textbf{CIFAR-10}} \\
 DBA        & 65.97 & 10.25 & 60.31 & 10.63 & 65.71 & 13.48 & 66.44 & 13.59 \\
 DBA+PPS    & 65.59 & 45.64 & 61.86 & 47.85 & 61.87 & 49.93 & 65.16 & 50.69 \\ 
 PPS+TOM    & 62.38 & \textbf{77.38} & 61.47 & \textbf{77.78} & 60.93 & \textbf{78.20} & 60.83 & \textbf{82.00} \\
\bottomrule
\end{tabular}
}
\caption{Ablation results of BAPFL on MNIST, FEMNIST, and CIFAR-10 datasets under different attack rates.}
\label{tab:ablation}
\end{table}

\subsection{Ablation Study}
\subsubsection{Effectiveness of BAPFL Components}
To evaluate the effectiveness of each component in BAPFL, we conduct an ablation study of BAPFL on three datasets, i.e., MNIST, FEMNIST, and CIFAR-10. Specifically, we examine the individual contributions of the PPS and TOM in BAPFL. The experimental results under different datasets and various attack rates are summarized in Table \ref{tab:ablation}. Across all datasets, BAPFL(PPS+TOM) consistently achieves the highest ASR with minimal impact on ACC. Removing either component significantly reduces ASR of BAPFL. For example, in the ablation study based on MNIST, when AR is 20\%, BAPFL(PPS+TOM) achieves 88.38\% ASR, while BAPFL(DBA+PPS) drops ASR to 69.43\%, and BAPFL(DBA) alone achieves only 42.44\%. Similar trends are observed in results based on FEMNIST and CIFAR-10. These results highlight that both PPS and TOM are essential for enhancing BAPFL’s effectiveness.

%Across all datasets, we observe that combining all three modules (TOM+PPS) consistently achieves the highest ASR with minimal impact on ACC, demonstrating the synergy between components. Notably, removing TOS or PPS significantly reduces the ASR. For example, on MNIST at 20\% attack rate, the BAPFL(TOS+PPS) method achieves 87.57\% ASR, while the variant without TOS (i.e., PFL+PPS) drops to 69.89\%, and PFL alone only reaches 64.91\%. Similar trends are observed on FEMNIST and CIFAR-10.

%These results confirm that both the prototype poisoning strategy and the trigger optimization module are critical to enhancing the effectiveness of BAPFL. In particular, the TOM module provides a significant boost in ASR by fine-tuning the trigger pattern to maximize its influence during personalized training.

% \begin{figure}[t]
%   \centering
%   \subfigure[Single target label]{
%     \includegraphics[width=0.47\linewidth]{proto-dynamic.pdf}
%   }
%   \subfigure[Round 160]{
%     \includegraphics[width=0.47\linewidth]{round160_client_batch1_trigger_result.png}
%   }
%   \caption{PCA visualization of benign and trigger prototypes of client 1 in different round. The trigger prototypes that are classified to the correct label $y$ are highlighted by circling.}
%   \label{fig:pca_evolution}
% \end{figure}

\begin{figure}[t]
  \centering
  \includegraphics[width=\linewidth]{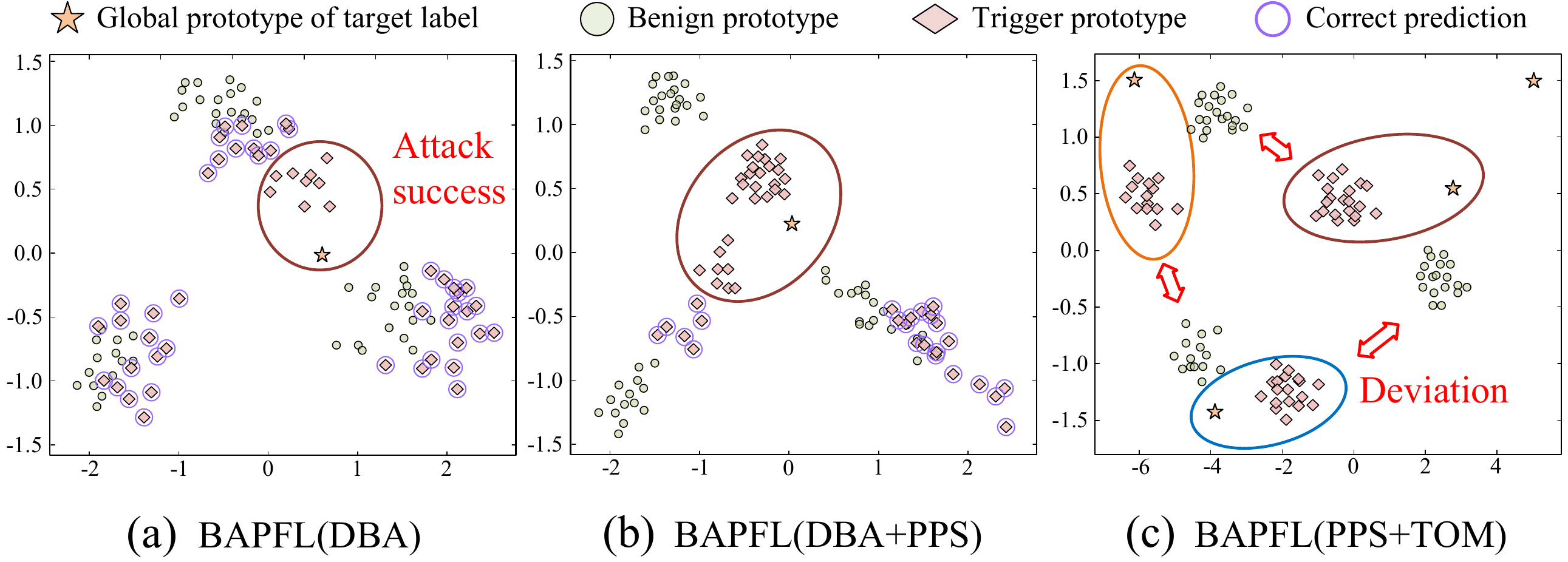}
  \caption{The principal component analysis (PCA) visualization of benign and trigger prototypes for benign client 1 under different attack strategies. The trigger prototypes classified as the original label are marked with purple circles.}
  \label{fig:pca_evolution}
\end{figure}

\subsubsection{Case Study}

To further illustrate the effectiveness of the PPS and TOM of BAPFL, we conduct a case study on FEMNIST, in which we visualize the prototype distribution and the classification results of trigger prototypes for a benign client under three attack strategies: BAPFL(DBA), BAPFL(DBA+PPS), and BAPFL(PPS+TOM). As shown in Figure~\ref{fig:pca_evolution}(a), the trigger prototypes of BAPFL(DBA) are optimized toward the global prototype of the target label, but most of them are still close to their corresponding benign prototypes. This leads to a low ASR. In contrast, in the case of BAPFL(DBA+PPS) (Figure~\ref{fig:pca_evolution}(b)), PPS separates the benign prototypes from the trigger prototypes. This enables more trigger prototypes to approach the global prototype of the target label, which increases the ASR. Finally, in the case of BAPFL(PPS+TOM) (Figure~\ref{fig:pca_evolution}(c)), TOM further expands the target label space and enhances the alignment of trigger prototypes with the global prototype of the target labels, achieving the highest ASR across all attack strategies. The above results indicate that both PPS and TOM in BAPFL play a crucial role in enhancing the ASR.

\section{Conclusions and Future Work}
%In this paper, we investigate the vulnerabilities to PFL to backdoor attacks and propose a novel and effective backdoor attack method BAPFL. 
In this paper, we investigate the problems of applying existing backdoor attacks in PFL and propose a novel and effective backdoor attack method BAPFL. By carefully designing poisoned prototypes and optimizing specific triggers for target labels, BAPFL successfully induces targeted misclassifications in benign models while evading detection. Comprehensive evaluations across diverse datasets and PFL frameworks demonstrate that BAPFL significantly improves ASR with negligible performance degradation on main tasks. BAPFL underscores the need for stronger defenses in PFL and provides insights into designing secure and trustworthy PFL systems. In future work, we intend to extend our methodology to other FL frameworks that adopt non-gradient-based aggregation strategies. 

%we plan to extend BAPFL to a wider range of FL scenarios, such as personalized federated learning, Knowledge transfer between large and small models, and federated graph learning. 

\bibliography{aaai2026}

%\input{7_ReproducibilityChecklist}
%\clearpage

\appendix
\section{Appendix A: Theoretical Analysis of BAPFL}
In this appendix, we formally analyze how the proposed BAPFL enhances backdoor effectiveness in prototype-based federated learning (PFL).

\subsection{Effectiveness of PPS}
We first introduce a key assumption that underlies prototype-based classification in PFL.

\textbf{Assumption 1.} In a well-trained PFL model, a sample $x$ is classified as label $k$ if its extracted feature $\phi(x)$ is closer (in $\ell_2$ distance) to the global prototype $\bar{P}^{(k)}$ than to any other prototype. That is,
\[
\text{label}(x) = \arg\min_k \|\phi(x) - \bar{P}^{(k)}\|_2.
\]
Under this assumption, the attack success rate (ASR) is increased if the features of trigger-embedded samples $\phi(T_{y_t}(x))$ are misaligned with the global prototype $\bar{P}^{(y)}$ of the original label and become closer to other class prototypes. We now show that PPS increases this misalignment by manipulating the process of prototype aggregation.

\textbf{Theorem 1.} The PPS increases the misclassification probability of trigger-embedded samples in benign models by poisoning the global prototype aggregation. Specifically, it manipulates the global prototype $\bar{P}^{(k)}$ of class $k$ to deviate from the trigger prototype $P_{tr}^{(k)}$, thereby misleading the optimization of benign prototypes and increasing the distance between the benign prototype $P_c^{(k)}$ and $P_{tr}^{(k)}$.

\textit{Proof:} $\bar{P}^{(k)}$ is computed as the average of clients' local prototypes, i.e.,
\[
\bar{P}^{(k)} = \frac{1}{C} \sum_{c=1}^C P_c^{(k)}.
\]
The malicious client $c^*$ uploads a poisoned prototype $P_{c^*}^{(k)}$ defined as:
\[
P_{c^*}^{(k)} = 2 \cdot P_{proj} - P_{tr}^{(k)},
\]
where 
\[
P_{proj} = \frac{\bar{P}^{(k)} \cdot P_{tr}^{(k)}}{\bar{P}^{(k)} \cdot \bar{P}^{(k)}} \cdot \bar{P}^{(k)}.
\]
This poisoned prototype $P_{c^*}^{(k)}$ is the reflection of $P_{tr}^{(k)}$ with respect to the projection point $P_{proj}$ on $\bar{P}^{(k)}$, thus intentionally pushing the aggregated prototype away from the direction of $P_{tr}^{(k)}$. Let $C_b$ denote the set of benign clients and $C_m$ the set of malicious clients. The new aggregated global prototype becomes:
\[
\bar{P}_{\text{new}}^{(k)} = \frac{1}{|C_b| + |C_m|} \left( \sum_{c \in C_b} P_c^{(k)} + \sum_{c^* \in C_m} P_{c^*}^{(k)} \right).
\]
Since the poisoned prototypes are reflected points away from $P_{tr}^{(k)}$, the vector $\bar{P}_{\text{new}}^{(k)} - P_{tr}^{(k)}$ increases in magnitude compared to $\bar{P}^{(k)} - P_{tr}^{(k)}$, i.e.,
\[
\|\bar{P}_{\text{new}}^{(k)} - P_{tr}^{(k)}\|_2 > \|\bar{P}^{(k)} - P_{tr}^{(k)}\|_2.
\]
In the subsequent training rounds, benign clients optimize their local prototypes $P_c^{(k)}$ to minimize the consistency loss $\mathcal{L}_P$ with the (now biased) global prototype:
\[
\mathcal{L}_{P} = \|P_c^{(k)} - \bar{P}_{\text{new}}^{(k)}\|_2.
\]
Thus, $P_c^{(k)}$ is continuously pulled toward $\bar{P}_{\text{new}}^{(k)}$, and consequently, $\|P_c^{(k)} - P_{tr}^{(k)}\|_2$ increases over training rounds. According to \textit{Assumption 1}, for a clean sample $x$, if its feature $\phi(x)$ approximates $P_{c}^{(k)}$, then the classification result of this sample $x$ is:
\[
k = \arg\min_j \|\phi(x) - \bar{P}_{\text{new}}^{(j)}\|_2 .
\]
For the corresponding trigger-embedded sample $x^{*}$ with target label $y_t$, its feature $\phi(x^{*})$ approximates $P_{tr}^{(k)}$. Since $\|P_c^{(k)} - P_{tr}^{(k)}\|$ becomes larger due to PPS, the probability that $\phi(x^{*})$ is closest to $\bar{P}_{\text{new}}^{(k)}$ decreases, i.e.,
\[
\Pr\left[ \arg\min_j \|\phi(x^{*}) - \bar{P}_{\text{new}}^{(j)}\|_2 = k \right] \downarrow.
\]
Therefore, PPS increases the misclassification probability of trigger-embedded samples in benign models.

\subsection{Effectiveness of TOM}
We now analyze how TOM increases the ASR in PFL. Specifically, by aligning trigger prototypes with the global prototypes of target labels that overlap with local label spaces, TOM increases the probability that the trigger-target label mapping is unintentionally activated in benign clients.
%TOM increases the probability that the trigger-target label mapping is unintentionally activated in benign clients by optimizing triggers aligned with the global prototype of the target labels that overlap with local label spaces.

\textbf{Assumption 2.} For a trigger-embedded sample $x^*$ with target label $y_t$, its classification is determined by the proximity of its feature $\phi(T_{y_t}(x))$ to the global prototype $\bar{P}^{(k)}$:
\[
\text{label}(T_{y_t}(x)) = \arg\min_k \|\phi(T_{y_t}(x)) - \bar{P}^{(k)}\|_2.
\]

\textbf{Theorem 2.} TOM increases the probability that the trigger-target label mapping is unintentionally activated by benign models. %TOM achieves this by expanding the target label set $Y_t$ to increase overlap with benign local label spaces and optimizing trigger embeddings such that the corresponding features align with the global prototypes of these target labels.

\textit{Proof:} Let $Y_t$ denote the set of target labels chosen by the attacker, where:
\[
Y_t = \bigcup_{c \in C_b} \mathcal{Y}_c,
\]
and $\mathcal{Y}_c$ is the local label space of benign client $c$. $Y_t$ maximizes the probability that any benign client $c$ has $y_t \in \mathcal{Y}_c$ for some $y_t \in Y_t$. For each target label $y_t \in Y_t$, TOM optimizes a dedicated trigger pattern $(\delta_{y_t}, M_{y_t})$ to construct a trigger function:
\[
T_{y_t}(x) = (1 - M_{y_t}) \odot x + M_{y_t} \odot \delta_{y_t}.
\]
TOM jointly optimizes $(\delta_{y_t}, M_{y_t})$ to minimize the loss $\mathcal{L}_{\text{trigger}}$. After training, $P_{tr}^{(y_t)} \approx \bar{P}^{(y_t)}$. Now consider a benign client $c$ such that $y_t \in \mathcal{Y}_c$. Suppose the trigger-embedded sample $T_{y_t}(x)$ is injected into $c$’s testing batch, TOM increases the likelihood that:
\[
\phi(T_{y_t}(x)) \approx P_{tr}^{(y_t)} \approx \bar{P}^{(y_t)}.
\]
Therefore, under \textit{Assumption 2}, the probability that a benign model classifies $T_{y_t}(x)$ as $y_t$ increases:
\[
\Pr\left[ \arg\min_k \|\phi(T_{y_t}(x)) - \bar{P}^{(k)}\|_2 = y_t \right] \uparrow.
\]
Hence, TOM increases the chance of “unintentional backdoor activation” across benign clients.

In conclusion, our theoretical analysis demonstrates that the integration of PPS and TOM enables BAPFL to effectively enhance the the ASR in PFL.

\begin{table}[t]
\centering
\resizebox{\linewidth}{!}{  % 缩放开始
\begin{tabular}{l|cc|cc|cc|cc}
\toprule
\multirow{2}{*}{Strategy} & \multicolumn{2}{c|}{10\%} & \multicolumn{2}{c|}{20\%} & \multicolumn{2}{c|}{30\%} & \multicolumn{2}{c}{40\%} \\
\cmidrule{2-9}
 & ACC & ASR & ACC & ASR & ACC & ASR & ACC & ASR \\
\midrule
\multicolumn{9}{c}{\textbf{MNIST}}\\
OBF & 98.10 & 87.10 & 97.52 & 87.58 & 96.67 & 88.18 & 96.79 & 89.40 \\
GPF & 97.71 & 77.18 & 98.15 & 83.23 & 98.21 & 85.67 & 98.01 & 87.57 \\
PFS & 97.96 & \textbf{87.14} & 97.85 & \textbf{88.38} & 96.89 & \textbf{88.89} & 96.90 & \textbf{91.08} \\

\midrule
\multicolumn{9}{c}{\textbf{FEMNIST}}\\
OBF & 91.69 & 83.45 & 91.62 & 85.36 & 90.49 & 86.60 & 89.90 & 87.56 \\
GPF & 91.11 & 81.88 & 91.06 & 83.02 & 89.93 & 84.23 & 89.56 & 85.82 \\
PFS & 91.94 & \textbf{87.39} & 91.29 & \textbf{88.48} & 90.55 & \textbf{89.19} & 89.18 & \textbf{89.23} \\

\midrule
\multicolumn{9}{c}{\textbf{CIFAR-10}}\\
OBF & 62.21 & 73.04 & 60.23 & 73.19 & 58.57 & 73.21 & 64.67 & 73.43 \\
GPF & 63.54 & 72.86 & 61.45 & 73.33 & 58.36 & 73.48 & 61.06 & 73.56 \\
PFS & 62.38 & \textbf{77.38} & 61.47 & \textbf{77.78} & 60.93 & \textbf{78.20} & 60.83 & \textbf{82.00} \\

\bottomrule
\end{tabular}
}
\caption{Performance comparison of BAFPL with different flipping strategies.}\label{tab:pps-com}
\end{table}

\section*{Appendix B: Comparison of Different \\ Flipping Strategies}

To validate the effectiveness of our proposed prototype flipping strategy (PFS) in the PPS, we compare it against two intuitive baselines: 1) \textit{Origin-based flipping (OBF)}. This strategy reflects the trigger prototype $P_{tr}^{(k)}$ based on the origin to construct the poisoned prototype $P_{c^*}^{(k)}$, i.e.,
\[
P_{c^*}^{(k)} = -P_{tr}^{(k)}.
\]
Although simple and effective, OBF reduces the stealthiness of the attack since it often yields unstable or easily detectable poisoned prototypes that are significantly different from the benign prototypes. 2) \textit{Global prototype-based flipping (GPF)}. This strategy reflects $P_{tr}^{(k)}$ with respect to the global prototype $\bar{P}^{(k)}$ to construct $P_{c^*}^{(k)}$, i.e.,
\[
P_{c^*}^{(k)} = 2 \cdot \bar{P}^{(k)} - P_{tr}^{(k)}.
\]
Compared with OBF, GPF achieves finer control over the direction of $P_{c^*}^{(k)}$, but it lacks control over the norm of $P_{c^*}^{(k)}$, potentially weakening attack effectiveness or introducing excessive perturbation.

Conversely, our PFS enables fine-grained control over both the direction and norm of $P_{c^*}^{(k)}$, achieving more precise manipulation of the global prototype while preserving stealth.

\textbf{Experimental Comparison.} In PFL, we report the main task accuracy (ACC) and ASR of BAPFL with different flipping strategies across three datasets (MNIST, FEMNIST, CIFAR-10) and varying attack rates (AR = 10\% to 40\%). The results are shown in Table \ref{tab:pps-com}. Across all datasets and attack rates, the BAPFL with PFS consistently achieves the highest ASR while maintaining comparable or even better ACC than other baselines. This demonstrates that our strategy provides a more effective and stealthy attack mechanism by precisely constructing the direction and norm of the poisoned prototypes.

\end{document}